\documentclass[abstract=true,a4paper]{scrartcl}

\usepackage{calc}
\usepackage[utf8]{inputenc}
\usepackage{tikz}
\usepackage{xcolor}
\usepackage{color}
\usepackage{natbib}
\usepackage{url}
\usepackage{booktabs}
\usepackage{pifont}

\newcommand{\xmark}{\ding{53}}%

\def\signed #1{{\leavevmode\unskip\nobreak\hfil\penalty50\hskip2em
  \hbox{}\nobreak\hfil(#1)%
  \parfillskip=0pt \finalhyphendemerits=0 \endgraf}}

\newsavebox\mybox

\title{Explainable Machine Learning\\ for Scientific Insights and Discoveries}
\author{Ribana Roscher\thanks{R. Roscher and J. Garcke contributed equally to this work} \textsuperscript{1,2}, Bastian Bohn\textsuperscript{3},\\ Marco F. Duarte\textsuperscript{4}, and Jochen Garcke\footnotemark[1] \textsuperscript{3,5}}
\publishers{\small \textsuperscript{1}Institute of Geodesy and Geoinformation, University of Bonn, Germany\\
\textsuperscript{2}Institute of Computer Science, University of Osnabrueck, Germany\\
\textsuperscript{3}Institute for Numerical Simulation, University of Bonn, Germany\\
\textsuperscript{4}Dept. of Electrical and Computer Engineering, University of Massachusetts Amherst, USA\\
\textsuperscript{5}Fraunhofer Center for Machine Learning and Fraunhofer SCAI, Sankt Augustin, Germany}

\date{}

\usepackage{natbib}
\usepackage{graphicx}

\begin{document}
\maketitle

\begin{abstract}
Machine learning methods have been remarkably successful for a wide range of application areas in the extraction of essential information from data.
An exciting and relatively recent development is the uptake of machine learning in the natural sciences, where the major goal is to obtain novel scientific insights and discoveries from observational or simulated data.
A prerequisite for obtaining a scientific outcome is domain knowledge, which is needed to gain explainability, but also to enhance scientific consistency.
In this article we review explainable machine learning in view of applications in the natural sciences and discuss three core elements which we identified as relevant in this context: \emph{transparency}, \emph{interpretability}, and \emph{explainability}.
With respect to these core elements, we provide a survey of recent scientific works that incorporate machine learning and the way that explainable machine learning is used in combination with domain knowledge from the application areas. 
   
\end{abstract}

\section{Introduction}
Machine learning methods, especially with the rise of neural networks (NNs), are nowadays used widely in commercial applications. 
This success has also led to a considerable uptake of machine learning (ML) in many scientific areas.  
Usually these models are trained with regard to high accuracy, but recently there is also a high demand for understanding the way a specific model operates and the underlying reasons for the produced decisions.
One motivation behind this is that scientists increasingly adopt ML for optimizing and producing scientific outcomes, where explainability is a prerequisite to ensure the scientific value of the outcome.
In this context, research directions such as explainable artificial intelligence (AI) \citep{SamITU18}, informed ML \citep{Rueden.ea.2019}, or intelligible intelligence \citep{weld2018challenge} have emerged. 
Though related, the concepts, goals, and motivations vary, and core technical terms are defined in different ways.

In the natural sciences, the main goals for utilizing ML are scientific understanding, inferring causal relationships from observational data, or even achieving new scientific insights. With ML approaches, one can nowadays (semi-)automatically process and analyze large amounts of scientific data from experiments, observations, or other sources. The specific aim and scientific outcome representation will depend on the researchers' intentions, purposes, objectives, contextual standards of accuracy, and intended audiences. Regarding conditions on an adequate scientific representation, we refer to the philosophy of science~\citep{Frigg.Nguyen:2018}.

This article provides a survey of recent ML approaches which are meant to derive scientific outcomes, where we specifically focus on the natural sciences. Given the scientific outcomes, novel insights can be derived helping for a deeper understanding, or scientific discoveries can be revealed which were not known before.
\emph{Gaining scientific insights and discoveries} from an ML algorithm means gathering information from its output and/or its parameters regarding the scientific process or experiments underlying the data.

One should note that a data-driven effort of scientific discovery is nothing new, but mimics the revolutionary work of Johannes Kepler and Sir Isaac Newton, which was based on a combination of data-driven and analytical work. As stated by \cite{Brunton.Kutz:2019}, 
\begin{quote}
Data science is not replacing mathematical physics and engineering, but is instead augmenting it for the twenty-first century, resulting in more of a renaissance than a revolution.
\end{quote}
What is new is the abundance of high-quality data in the combination with scalable computational and data processing infrastructure.

\begin{figure}[ht]
    \centering
    \includegraphics[width=0.9\textwidth]{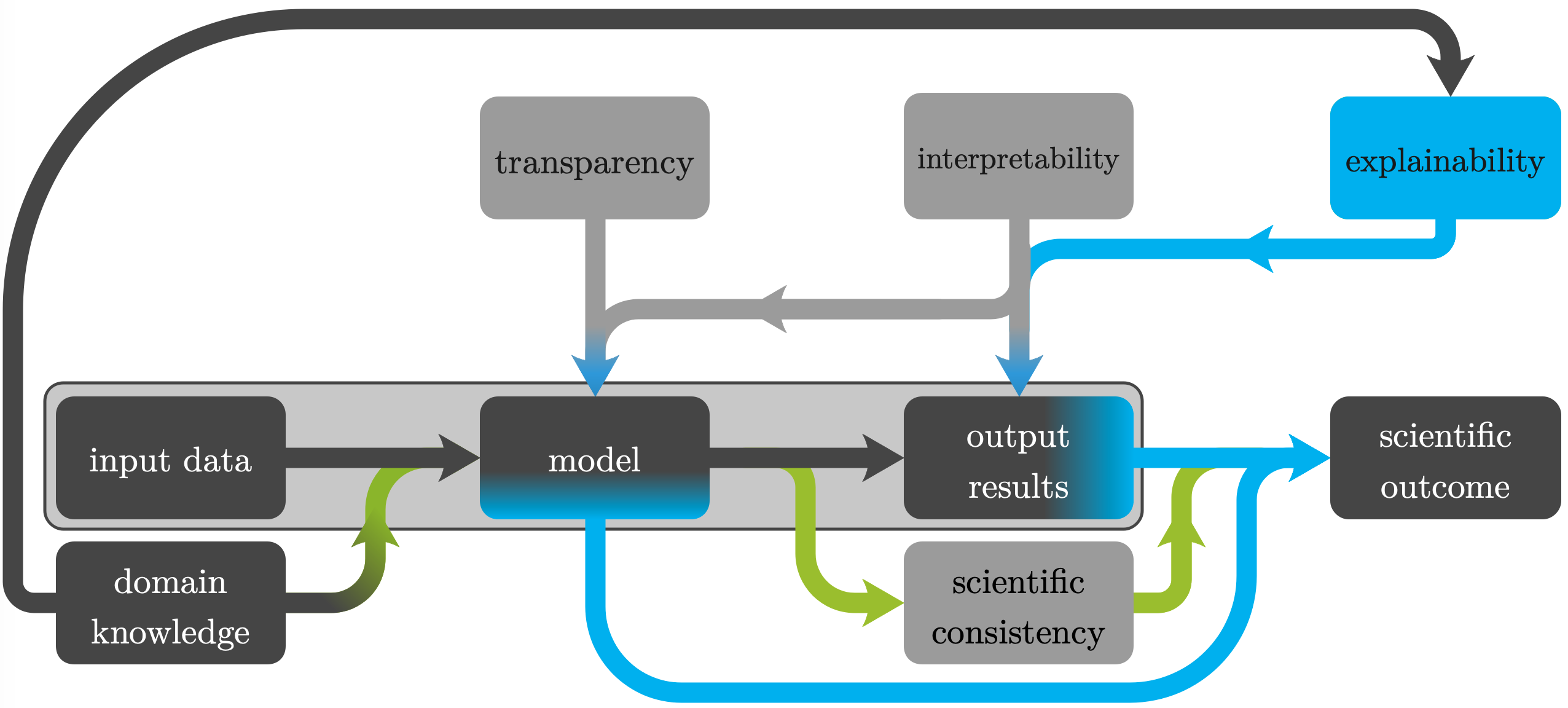}
    \caption{Major ML-based chains from which scientific outcomes can be derived: The commonly used, basic ML chain (light gray box) learns a black box model from given input data and provides an output. Given the black box model and input-output relations, a scientific outcome can be derived by explaining the output results utilizing domain knowledge. Alternatively, a transparent and interpretable model can be explained using domain knowledge leading to scientific outcomes. Additionally, the incorporation of domain knowledge can promote scientifically consistent solutions (green arrows).}
    \label{fig:mindmap}
\end{figure}

The main contribution of this survey is the discussion of commonly used ML-based chains leading to scientific outcomes which have been used in the natural sciences (see Fig.~\ref{fig:mindmap}).
A central role play the three elements \emph{transparency}, \emph{interpretability}, and \emph{explainability}, which will be defined and discussed in detail in this survey. 
The core is the basic ML chain, in which a model is learned from given input data and with a specific learning paradigm, yielding output results utilizing the learned model.
In order to derive a scientific outcome, either the output results or the model is explained, where interpretability is the prerequisite for explainability.
Moreover, transparency is required to explain a model. 
A further essential part is \emph{domain knowledge}, which is necessary to achieve explainability, but can also be used to foster \emph{scientific consistency} of the model and the result.
Generally, providing domain knowledge to an algorithm means to enhance the input data, model, optimizer, output results, or any other part of the ML algorithm by using information gained from domain insights such as laws of nature and chemical, biological, or physical models~\citep{Rueden.ea.2019}.
Besides the purpose of explainability, integrating domain knowledge can help with model tractability and regularization in scenarios where not enough data is available. It might also increase the performance of a model or reduce computational time. 
We will give diverse examples from the natural sciences of approaches that can be related to these topics.
Moreover, we define several groups based on the presence of the components of the ML chains from Fig.~\ref{fig:mindmap} and assign each example to one group.
Our goal is to foster a better understanding and a clearer overview of ML algorithms applied to data from the natural sciences.

The paper is structured as follows.
In Section~\ref{sec:terminology} we discuss transparency, interpretability, and explainability in the context of this article.
While these terms are more methodology-driven and refer to properties of the model and the algorithm, we also describe the role of additional information and domain knowledge, as well as scientific consistency.
In Section~\ref{sec:soml}, we highlight several applications in the natural sciences that use the mentioned concepts to gain new scientific insights, while organizing the ML workflows into characteristic groups based on the different uptakes of interpretability and explainability.

\section{Terminology}
\label{sec:terminology}

It can be observed that in the literature about explainable ML several descriptive terms are used with diverse meanings, see e.g.~\cite{Doshi-Velez2017,Gilpin2018,Guidotti2018,Lipton2018,Montavon2018,Murdoch2019}. Nonetheless, distinct ideas can be identified. %~\citep{Lipton2018}. 
For the purpose of this work, we differentiate between \emph{transparency}, \emph{interpretability}, and \emph{explainability}. Roughly speaking, transparency considers the ML approach, interpretability considers the ML model together with data, and explainability considers the model, the data, and human involvement. 

\paragraph{Transparency}
An ML approach is transparent if the processes that extract model parameters from training data and generate labels from testing data can be described and motivated by the approach designer. We say that the transparency of an ML approach concerns its different ingredients: this includes the overall model structure, the individual model components, the learning algorithm, and how the specific solution is obtained by the algorithm. We propose to differentiate between \emph{model transparency}, \emph{design transparency}, and \emph{algorithmic transparency}.  Generally, to expect an ML method to be completely transparent in all aspects is rather unrealistic; usually there will be different degrees of transparency.

As an example, consider kernel-based ML approaches~\citep{Hofmann2008,Rasmussen.Williams:2006}. 
The obtained model is transparent as it is given as a sum of kernel functions. The individual design component is the chosen kernel. Choosing between a linear or nonlinear kernel is typically a transparent design decision. However, using the common Gaussian kernel based on Euclidean distances can be a non-transparent design decision. In other words, it may not be clear why a given nonlinear kernel was chosen. Domain specific design choices can be made, in particular using suitable distance measures to replace the Euclidean distance, making the design of this model component (more) transparent. 
In the case of GP regression, the specific choice of the kernel can be built into the optimization of the hyper-parameters using the maximum likelihood framework~\citep{Rasmussen.Williams:2006}. Thereby, design transparency goes over to algorithmic transparency.
Furthermore, the obtained specific solution is, from a mathematical point of view, transparent. Namely, it is the unique solution of a convex optimization problem which can be reproducibly  obtained~\citep{Hofmann2008,Rasmussen.Williams:2006}, resulting in algorithmic transparency. In contrast,  approximations in the specific solution method such as early stopping, matrix approximations, stochastic gradient descent, and others, can result in (some) non-transparency of the algorithm.

As another example, consider NNs~\citep{Goodfellow-et-al-2016}. 
The model is transparent since its input-output relation and structure can be written down in mathematical terms.
Individual model components, such as a layer of a NN, that are chosen based on domain knowledge can be considered as design transparent. Nonetheless, the layer parameters --- be it their numbers, size, or nonlinearities involved --- are often chosen in an ad-hoc or heuristic fashion and not motivated by knowledge; these decisions are therefore not design transparent. The learning algorithm is typically transparent, e.g., stochastic gradient descent can be easily written down. However, the choice of hyper-parameters such as learning rate, batch size, etc., has more a heuristic, non-transparent algorithmic nature. Due to the presence of several local minima, the solution is usually not easily reproducible; therefore, the obtained specific solution is not (fully) algorithmically transparent. 

Our view is closely related with \cite{Lipton2018}, who writes:
\begin{quote}
Informally, transparency is the opposite of opacity or
“black-boxness.” It connotes some sense of understanding
the mechanism by which the model works. Transparency
is considered here at the level of the entire model
(simulatability), at the level of individual components such
as parameters (decomposability), and at the level of the
training algorithm (algorithmic transparency).
\end{quote}

An important contribution to the understanding of ML algorithms is their mathematical interpretation and derivation, which help to understand when and how to use these approaches. 
Classical examples are the Kalman filter or principal component analysis, where several mathematical derivations exist for each and enhance their understanding. 
Note that although there are many mathematical attempts to a better understanding of deep learning, at this stage ``the [mathematical] interpretation of NNs appears to mimic a type of Rorschach test,'' according to~\cite{Charles2018}. 

Overall, we argue that transparency in its three forms does to a large degree not depend on the specific data, but solely on the ML method. But clearly, the obtained specific solution, in particular the ``solution path'' to it by the (iterative) algorithm, depends on the training data. The analysis task and the type of attributes usually play a role in achieving design transparency.
Moreover, the choice of hyper-parameters might involve model structure, components, or the algorithm, while in an algorithmic determination of hyper-parameters the specific training data comes into play again. 

\paragraph{Interpretability}
We consider interpretability as about making sense of the obtained ML model. Generally, to interpret means ``to explain the meaning of'' or ``present in understandable terms''\footnote{https://www.merriam-webster.com/dictionary/interpret}; see also~\cite{Doshi-Velez2017,Gilpin2018,Guidotti2018}. 
We consider explaining as a separate aspect, on top of an interpretation, and focus here on the second aspect.
Therefore, the aim of interpretability is to present some of the properties of an ML model in understandable terms to a human. Ideally, one could answer the question from \cite{Casert2018}: ``Can we understand on what the ML algorithm bases its decision?''
Somewhat formally, \cite{Montavon2018} state:
\begin{quote}
An interpretation is the mapping of an abstract concept (e.g., a predicted class) into a domain that the human can make sense of.
\end{quote}

Interpretations can be obtained by way of understandable proxy models, which approximate the predictions of a more complex approach~\citep{Gilpin2018,Guidotti2018}. Longstanding approaches involve decision trees or rule extraction~\citep{Andrews1995} and linear models. In prototype selection, one or several examples similar to the inspected datum are selected, from which criteria for the outcome can be obtained. 
For feature importance, the weights in a linear model are employed to identify attributes which are relevant for a prediction, either globally or locally.
For example, \cite{Ribeiro2016} introduced the model-agnostic approach LIME (Local Interpretable Model-Agnostic Explanations), which gives interpretation by creating locally a linear proxy model in the neighborhood of a datum, while the scores in layer-wise relevance propagation (LRP) are obtained by means of a first-order Taylor expansion of the nonlinear function~\citep{Montavon2018}.
Sensitivity analysis can be used to inspect how a model output (locally) depends upon the different input parameters~\citep{Saltelli2004}. 
Such an extraction of information from the input and the output of a learned model is also called \emph{post hoc interpretability}~\citep{Lipton2018} or \emph{reverse engineering}~\citep{Guidotti2018}.
Further details, types of interpretation, and specific realization can be found in recent surveys~\citep{Adadi2018,Gilpin2018,Guidotti2018}.

Visual approaches such as saliency masks or heatmaps show relevant patterns in the input based on feature importance,  sensitivity analysis, or relevance scores to explain model decisions, in particular employed for deep learning approaches for image classification~\citep{Hohman2018,Montavon2018,olah2018the}. 
Note that a formal notion for interpreting NNs was introduced, where a set of input features is deemed relevant for a classification decision if the expected classifier score remains nearly constant when randomising the remaining features~\citep{Macdonald.Kutyniok:2019}. The authors prove that under this notion the problem of finding small sets of relevant features is NP-hard, even when considering approximation within any non-trivial factor. This shows on the one hand the difficulty of algorithmically determining interpretations, and on the other hand justifies the current use of heuristic methods in practical applications.

In unsupervised learning, the analysis goal can be a better understanding of the data. For an example, by an interpretation of the obtained representation by linear or nonlinear dimensionality reduction~\citep{Lee.Verleysen:2007,Cichocki.Zdunek.ea:2009}, or by inspecting the components of  a low-rank tensor decomposition~\citep{Morup2011}.

Note that, in contrast to transparency, to achieve interpretability the data is always involved. Although there are model-agnostic approaches for interpretability, transparency or retaining the model can assist in the interpretation. %TODO cite
Furthermore, method specific approaches depend on transparency, for example layer-wise relevance propagation for NNs exploits the known model layout~\citep{Montavon2018}.

While the methods for interpretation allow the inspection of a single datum, \cite{Lapuschkin2019} observe that it quickly becomes very time consuming to investigate large numbers of individual interpretations.
As a step to automate the processing of the individual interpretations for a single datum, they employ clustering of heatmaps of many data to obtain an overall impression of the interpretations for the predictions of the ML algorithm. 

Finally, note that the interpretable and human level understanding of the performance of an ML approach can result in a different choice of the ML model, algorithm, or data pre-processing later on. 

\paragraph{Explainability}

While research into explainable ML is widely recognized as important, a joint understanding of the concept of explainability still needs to evolve. Concerning explanations, it has also been argued that there is a gap of expectations between ML and so-called explanation sciences such as law, cognitive science, philosophy, and the social sciences~\citep{Mittelstadt2018}. 

While in philosophy and psychology explanations have been the focus for a long time, a concise definition is not available. For example, explanations can differ in  completeness or the degree of causality. 
We suggest to follow a model from a recent review relating insights from the social sciences to explanations in AI~\citep{Miller2019}, which places explanatory questions into three classes: (1) what--questions, such as ``What event happened?''; (2) how--questions, such as ``How did that event happen?''; and (3) why--questions, such as ``Why did that event happen?''. 
From the field of explainable AI we consider a definition from \cite{Montavon2018}:
\begin{quote}
An explanation is the collection of features of the interpretable domain, that have contributed for a given example to produce a decision (e.g., classification or regression).
\end{quote}
As written in \cite{Guidotti2018}, ``[in explainable ML] these definitions assume implicitly that the concepts expressed in the understandable terms composing an explanation are self-contained and do not need further explanations.'' 

We believe on the other hand, that a collection of interpretations can be an explanation only with further contextual information, stemming from domain knowledge and related to the analysis goal. In other words, explainability usually cannot be achieved purely algorithmically. 
On its own, the interpretation of a model --- in understandable terms to a human --- for an individual datum might not provide an explanation to understand the decision. For example, the most relevant variables might be the same for several data, but the important observation for an understanding of the overall predictive behavior could be that in a ranking with respect to the interpretation, different variable lists are determined for each data as being of relevance. Overall, the result will depend on the underlying analysis goal. ``Why is the decision made?'' will need a different explanation than ``Why is the decision for datum A different to (the nearby) datum B?''.

In other words, for explainability, the goal of the ML ``user'' is very relevant. 
According to \cite{Adadi2018}, there are essentially four reasons to seek explanations: to justify decisions, to (enhance) control, to improve models, and to discover new knowledge.
For regulatory purposes it might be fine to have an explanation by examples or (local) feature analysis, so that certain ``formal'' aspects can be checked.
But, to attain scientific outcomes with ML one wants an understanding. 
Here, the scientist is using the data, the transparency of the method, and its interpretation to explain the output results (or the data) using domain knowledge and thereby to obtain a scientific outcome.

Furthermore, we suggest to differentiate between \emph{scientific explanations} and \emph{algorithmic explanations}.  
For scientific explanations, \cite{Overton2013} identifies five broad categories to classify the large majority of objects that are explained in science: data, entities, kinds, models, and theories. Furthermore, it is observed that whether there is a unifying general account of scientific explanation remains an open question. With an algorithmic explanation, one aims to reveal underlying causes to the decision of an ML method; this is what explainable ML aims to address.
In recent years, a focus on applying interpretation tools to better explain the output of a ML model can be observed. This can be seen in contrast to 
symbolic AI techniques, e.g., expert or planning systems, which in contrast are often seen as explainable per se. Hybrid systems of both symbolic and, so-called, connectionist AI, e.g., artificial NNs, are investigated to combine advantages from both approaches.
For example, \cite{liao2017object} propose ``object-oriented deep learning'' with the goal to convert a NN to a symbolic description to gain interpretability and explainability.
They state that generally in NNs there is inherently no explicit representation of symbolic concepts like objects or events, but rather a feature-oriented representation, which is difficult to explain.
In their representation, objects could be formulated to have disentangled and interpretable properties.
Although not commonly used so far, their work is one example for a promising direction towards a higher explainability of models. 

In the broader context, other properties that can be relevant when considering explainability of ML algorithms are safety/trust, accountability, reproducibility, transferability, robustness and multi-objective trade-off or mismatched objectives, see e.g.~\citep{Doshi-Velez2017,Lipton2018}. 
For example, in societal contexts reasons for a decision often matter. 
Typical examples are (semi-)automatic loan applications, hiring decisions, or risk assessment for insurance applicants, where one wants to know why a model gives a certain prediction and how one might be affected by those decisions.
In this context, and also due to regulatory reasons, one goal is that decisions based on ML models involve a fair and ethical decision making.
The importance to give reasons for decisions of an ML algorithm is also high for medical applications, where a motivation is the provision of trust in decisions such that patients are comfortable with the decision made.
All this is supported by the General Data Protection Regulation, which contains new rules regarding the use of personal information. 
One component of these rules can be summed up by the phrase ``right to an explanation''~\citep{Goodman2017}. 
Finally, for ML models deployed for decision-support and automation, in particular in potentially changing environments, an underlying assumption is that robustness and reliability can be better understood, or easier realized, if the model is interpretable~\citep{Lipton2018}.

One should also observe that explanations can be used to manipulate. For illustration, \cite{Baumeister1994} distinguish between the intuitive scientist, who seeks  to make the most accurate or otherwise optimal decision, and the intuitive lawyer, who desires to justify a preselected conclusion. 
With that in mind, one often aims for human-centric explanations of black-box models. There are simple or purely algorithmic explanations, e.g., based on emphasising relevant pixels in an image. In so-called slow judgements tasks, an explanation might more easily enforce confirmation biases. For example, using human-centric explanations as evaluation baselines can be biased towards certain individuals. Further,
a review of studies of experimental manipulations that require people to generate explanations or imagine scenarios indicates that people express greater confidence in a possibility, although false, when asked to generate explanations for it or imagine the possibility~\citep{Koehler1991}. 

\paragraph{Domain knowledge}
As outlined, domain knowledge is an essential part of explainability, but also for treating small data scenarios or for performance reasons.
A taxonomy for the explicit integration of knowledge into the ML pipeline, dubbed \emph{informed ML}, is proposed by~\cite{Rueden.ea.2019}.
Three aspects are involved:
\begin{itemize}
    \item type of knowledge,
    \item representation and transformation of knowledge, and
    \item integration of knowledge into the ML approach.
\end{itemize}
See also the related works of~\cite{Karpatne2017}, who use the term \emph{theory-guided data science}, or \emph{physics-informed learning} by \cite{Raissi2017a}.
For the purpose of this article, we follow~\cite{Rueden.ea.2019}, who arrange different types of knowledge along their degree of formality, from the sciences, over (engineering or production) process flow to world knowledge and finally individual (expert’s) intuition.
Knowledge can be assigned to several of the types in this incomplete list.

In the sciences, knowledge is often given in terms of mathematical equations, such as analytic expressions or differential equations, or as relations between instances and/or classes in form of rules or constraints. Its representation can for example be in the form of ontologies,  symmetries, or similarity measures. Knowledge can also be exploited by numerical simulations of models or through human interaction. 

As ingredients of an ML approach one considers the training data, the hypothesis space, the training algorithm, and the final model. In each of these, one can incorporate additional knowledge. 
Feature engineering is a common and longstanding way to incorporate knowledge into the training data, while using numerical simulations to generate (additional) training data is a modern phenomena.
One common way to integrate knowledge into the hypothesis space is by choosing the structure of the model. For example, by defining a specific architecture of a NN or by choosing a structure of probability distributions which observes existing or non-existing links between variables.
An example for the training phase is modifying the loss function according to additional knowledge, for example by adding a consistency term. 
Finally, the obtained model can be put in relation to existing
knowledge, for example by checking known constraints for the predictions. 

\paragraph{Scientific consistency}
A fundamental prerequisite for generating reliable outcomes for scientific applications is scientific consistency.
This means that the result obtained is plausible and consistent with existing scientific principles. 
The selection and formulation of the scientific principles to be met is based on domain knowledge, where the manner of integration is the core research question in areas such as informed ML.
In the chain of Fig.~\ref{fig:mindmap}, scientific consistency can be considered a priori at the model design stage or a posteriori by analysing the output results.
As pointed out by \cite{Rueden.ea.2019}, scientific consistency at the design stage can be understood as the result of a regularization effect, where various ways exist to restrict the solution space to scientifically consistent solutions.
\cite{reichstein2019deep} identify scientific consistency besides interpretability as one of the five major challenges we need to tackle to successfully adopt deep learning approaches in the geosciences.
\cite{Karpatne2017} underlines the importance of consistency by defining it as an essential component to measure performance:
\begin{quote}
One of the overarching visions of [theory-guided data science] is to include [..] consistency as a critical component of model performance along with training accuracy and model complexity. This can be summarized in a simple way by the following revised objective of model performance [...]: Performance $\propto$ Accuracy + Simplicity + Consistency.
\end{quote}
They discuss several ways to restrict the solution space to physically
consistent solutions, e.g., by (1) designing the model family, such as
specific network architectures; (2) guiding a learning algorithm
using, e.g., specific initializations, constraints, or (loss) regularizations; (3) refining the model output, e.g., using closed-form equations or model simulations; (4) formulating hybrid models of theory and ML, and (5) augmenting theory-based models using real data such as data assimilation or calibration.

Overall, the explicit restriction of the solution space to scientifically consistent and plausible solutions is not a requirement to achieve valuable scientific outcomes. 
Neglecting this restriction, however, means that a consistent and plausible solution cannot be guaranteed, even if an optimal result has been achieved from a mathematical point of view.

\section{Scientific Outcomes From Machine Learning}
\label{sec:soml}

In this section, we review examples that use ML and strive for different levels of transparency, interpretability, or explainability to produce scientific outcomes. 
To structure the different ML chains we define common groups and describe representative papers for each.
In Table \ref{tab:group} we specify the four major groups and several subgroups in more detail. We expect that examples for additional subgroups can be found, but that will not affect our core observations.
In particular, we distinguish between the following components: 

\begin{description}
    \item[Transparency] 
    We consider a model as design-transparent if the model, or parts of it, was chosen for specific reasons, usually due to knowledge from the application domain. 
    We call a model algorithmically transparent if the determination of the solution is obvious and traceable. 
    In view of reproducible science it is not surprising that essentially all the examples we found can be considered to be model-transparent.  \item[Interpretability] 
    We have a closer look at two types of interpretability. First, we consider model components, such as neurons in a NN or obtained latent variables, to be interpretable if they are represented in a way that can be further explained, for example with domain knowledge. 
    Second, the scientific outcome, i.e., the decision of the model, can be interpreted by the input, for example by using heatmaps.
    \item[Integration of domain knowledge] We will look at several ways how domain knowledge can be integrated. On the one hand, domain knowledge is needed to explain --- either to explain the scientific outcome or to derive scientific findings from the model or individual model components. On the other hand, domain knowledge can be integrated to enforce scientifically plausible and consistent results. This can be done in different ways, see \cite{Rueden.ea.2019}. Besides the integration of domain knowledge during the learning process of the model, it can also be used for post-hoc checks, where the scientific plausibility and consistency of the results is checked and possibly invalid results are removed.
\end{description}

The following collection of research works is a non-exhaustive selection from the literature of the last few years, where we aim to cover a broad range of usages of ML with a variety of scientific outcomes.
Furthermore, we focus on examples which utilize an extensive amount of scientific domain knowledge from the natural sciences. 
Due to the recent uptake of NNs in the sciences these tend to be the dominating  employed ML approach in current literature. Nonetheless, many of the described ML workflows or the approaches to integrate domain knowledge can be performed with other ML methods as well. 
Note that the assignment of a given article to a group is not always a clear judgement, in particular in view how and where domain knowledge is employed and in what form, and to what extent, an explanation is derived.

\begin{table}[ht]
  \begin{center}
    \caption{Group 1 includes approaches without any means of interpretability. 
    In Group 2, a first level of interpretability is added by employing domain knowledge to design the models or explain the outcomes. Group 3 deals with specific tools included in the respective algorithms or applied to their outputs to make them interpretable. Finally, Group 4 lists approaches where scientific insights are gained by explaining the machine learning model itself.}
    \label{tab:group}
    \begin{tabular}{c|cc|cc|cccc} 
    \toprule
    Group & \multicolumn{2}{c}{Transparency} & \multicolumn{2}{c}{Interpretability} & \multicolumn{4}{c}{Integration of domain knowledge} \\
    & design & alg. & model & in-out & explaining & explaining & design & post-hoc \\
    &  &  &  &  & model & outcomes &  & check \\
      \midrule 
     1a  & - & - & - & -  & - & - & - & - \\
     1b & \xmark & - & -  & - & - & - & - & -\\
     1c & \xmark & - & -  & - & - & - & \xmark & -\\
     \midrule
     2a & \xmark & - & - & - & - & \xmark & \xmark & - \\
     2b & \xmark & \xmark & - & -  & - & \xmark & \xmark & -\\
     2c & \xmark & - & - & -  & - & \xmark & \xmark & \xmark \\
     \midrule
     3a & - & - & - & \xmark  & - & \xmark & - & - \\
     3b & \xmark & - & - & \xmark & - & \xmark & - & - \\
     3c & \xmark & - & - & \xmark  & - & \xmark & \xmark & - \\
     3d  & \xmark & - & \xmark & -  & - & \xmark & \xmark & - \\
     3e & \xmark & - & \xmark & -  & - & \xmark & \xmark & \xmark \\
     \midrule
     4a & \xmark & - & \xmark & -  & \xmark & \xmark & \xmark & - \\
      4b  & \xmark & \xmark & \xmark & -  & \xmark & \xmark & \xmark &  - \\
      \bottomrule
    \end{tabular}
  \end{center}
\end{table}

\subsection{Scientific Outcomes by Explaining Output Results}
\label{sec:inout}

Many works address the derivation of outcomes by learning an ML model and generalizing from known input-output relations to new input-output pairs.
This states the lowest degree of explainability with no necessity of a transparent or interpretable model.
In the case that a scientifically useful outcome is to be estimated, most of these approaches so far solely explain what the outcome is from a scientific point of view (scientific explanation), but cannot answer the question of {\em why this specific outcome was obtained} from an algorithmic point of view (algorithmic explanation). 
Other approaches attempt to scientifically explain the output in terms of the specific corresponding input, given a learned model.
Here, interpretation tools are utilized, where the model is used only as a means to an end to explain the result and it is not explicitly analyzed itself.

\subsubsection{Prediction of Intuitive Outcomes}
The derivation of intuitive physics is a task which is often considered in papers from the following group.
Intuitive physics are everyday-observed rules of nature which help us to predict the outcome of events even with a relatively untrained human perception \citep{mccloskey1983intuitive}.

\vspace{1em}
\emph{Group 1a (basic ML chain): A black-box approach, or at most model-transparent approach, is used to derive an outcome. It is not interpretable and cannot be explained. The outcome is not or only hardly explainable from a scientific point of view.}
\vspace{1em}

\cite{Lerer2016}, for example, use video simulations to learn intuitive physics, e.g., about the stability of wooden block towers.
They use ResNet-34 
and Googlenet 
and formulate a binary classification task to predict whether these towers will fall. 
In a similar way, but with more complex scenes or differently shaped objects, \cite{Li2016} predict the physical stability of stacked objects using various popular convolutional neural networks (CNNs) architectures.  
\cite{forster2019hyperspectral} predict the spread of diseases on barley plants in microscopic hyperspectral images by generating highly probable image-based appearances over the course of several days. 
They use cycle-consistent generative adversarial networks to learn how an image will change from one day to the next or to the previous day, however, without any biological parameters involved.
\cite{Mauro2016} present an approach for the design of new functional glasses which comprises the prediction of characteristics relevant for manufacturing as well as end-use properties of glass. 
They utilize NNs to estimate the liquidus temperatures for various silicate compositions consisting of up to eight different components. 
Generally, the identification of an optimized composition of the silicates yielding a suitable liquidus temperature is a costly task and is oftentimes based on trial-and-error. 
For this, they learn from several hundred composites with known output properties and apply the model to novel, unknown composites.
In their workflow, they also consider, outside of the ML chain, other quantities of interest which are derived by physics-driven models.
\cite{raissi2018deep} proposes a nonlinear regression approach employing NNs to learn closed form representations of partial differential equations (PDEs) from scattered data collected in space and time, thereby uncovering the dynamic dependencies and obtaining a model that can be subsequently used to forecast future states. In benchmark studies, including Burgers' equation, nonlinear Schrödinger equation, or Navier-Stokes equation, the underlying dynamics are learned from numerical simulation data up to a specific time. The obtained model is used to forecast future states, where relative $L_2$-errors of up to the order of $10^{-3}$ are observed.
While the method inherently models the PDEs and the dynamics themselves, the rather general  network model does not allow to draw direct scientific conclusions on the structure of the underlying process.

\vspace{1em}
\emph{Group 1b: These models are not only model- but also design-transparent to some extent, where the design process is influenced by knowledge in the specific domain.}
\vspace{1em}

Besides the simple prediction network presented by \cite{Lerer2016} in Group 1a, they also propose a network called PhysNet to predict the trajectory of the wooden blocks in case the tower is collapsing.
It is formulated as a mask prediction network trained for instance segmentation, where each wooden block is defined as one class.
The construction of PhysNet is made design-transparent in the sense that the network is constructed to capture the arrangement of blocks by using alternating upsampling and convolution layers, and an increased depth to reason about the block movement, as well.
PhysNet outperforms human subjects on synthetic data and achieves comparable results on real data.
\cite{ansdell2018scientific} designed a multi-source NN for exoplanet transit classification. 
They integrate additional information by adding identical information about centroid time-series data to all input sources, which is assumed to help the network learn important connections, and by concatenating the output of a hidden layer with stellar parameters, as it is assumed they are correlated with classification. 
\cite{zhu2015understanding} introduces a framework which calculates physical concepts from color-depth videos that estimates tool and tool-use such as cracking a nut. 
In their work, they learn task-oriented representations for each tool and task combination defined over a graph with spatial, temporal, and causal relations.
They distinguish between 13 physical concepts, e.g., painting a wall, and show that the framework is able to generalize from known to unseen concepts by selecting appropriate tools and tool-uses. 
A hybrid approach is presented by \cite{BJW18} to successfully model the properties of contaminant dispersion in soil. 
They extract temporal information from dynamic data using a long short-term memory network and combine it with static data in a NN.
In this way, the network models the spatial correlations underlying the dispersion model, which are independent of the location of the contaminant.
\cite{BL18} have proposed a data-centric approach for scientific design based on the combination of a generative model for the data being considered, e.g., an autoencoder trained on genomes or proteins, and a predictive model for a quantity or property of interest, e.g., disease indicators or protein fluorescence. 
For DNA sequence design, these two components are integrated by applying the predictive model to samples from the generative model. In this way, it is possible to generate new synthetic data samples that optimize the value of the quantity or property by leveraging an adaptive sampling technique over the generative model; see also \cite{pmlr-v97-brookes19a}.

\vspace{1em}
\emph{Group 1c: Here, in addition to group 1b, the design process is influenced by domain knowledge regarding the model, the cost function, or the feature generation process to enhance scientific consistency and plausibility.}
\vspace{1em}

\cite{Tompson2016} and \cite{jeong2015data} use physics-informed approaches for applications such as fluid simulations based on the incompressible Navier-Stokes equations, where physics-based losses are introduced to achieve plausible results. 
The idea in \cite{Tompson2016} is to use a transparent cost function design by reformulating the condition of divergence-free velocity fields into an unsupervised learning problem at each time step. 
The random forest model used in \cite{jeong2015data} to predict a fluid particle's velocity can be viewed as a transparent choice per se due to its simple nature.
\cite{hagensieker2017tropical} classify land use and land cover and their changes based on remote sensing satellite timeseries data. 
They integrate domain knowledge about the transition of specific land use and land cover classes such as forest or burnt areas to increase the classification accuracy.
They utilize a discriminative random field with transition matrices that contain the likelihoods about land cover and land use changes to enforce, for example, that a transition from burnt area to forest is highly unlikely.

\subsubsection{Prediction of Scientific Parameters and Properties}
Although the approaches just described set up prediction as a supervised learning problem, there is still a gap between common supervised tasks, e.g., classification, object detection, and prediction, and actual understanding of a scene and its reasoning. 
Like a layman in the corresponding scientific field, the methods presented so far do not learn a model that is able to capture and derive scientific properties and dynamics of phenomena or objects and their environment, as well as their interactions.
Therefore, the model cannot inherently explain why an outcome was obtained from a scientific viewpoint.
\cite{chang2017compositional} denote these respective approaches as bottom-up, where observations are directly mapped to an estimate of some behavior or some outcome of a scene.
To tackle the challenge of achieving a higher explainability and a better scientific usability, several so-called top-down classification and regression frameworks have been formulated which infer scientific parameters.
In both cases, only the scientific explanation is sought.

\vspace{1em}
\emph{Group 2a: In these ML models, domain knowledge is incorporated, often to enforce scientific consistency. Therefore, the design process is partly transparent and tailored to the application. The outcome is explainable from a scientific point of view since scientific parameters and properties are derived, which can be used for further processing.}
\vspace{1em}

\cite{Stewart2017}, for example, detect and track objects in videos in an unsupervised way.
For this, they use a regression CNN and introduce terms during training which measure the consistency of the output when compared to physical laws which specifically and thoroughly describe the dynamics in the video. 
In this case, the input of the regression network is a video sequence and the output is a time-series of physical parameters such as the height of a thrown object. 
By incorporating domain knowledge and image properties into their loss functions, part of their design process becomes transparent and explainability is gained due to comparisons to the underlying physical process. 
However, the model and algorithms are not completely transparent since standard CNNs with an ADAM minimizer are employed. Although this choice of model and algorithm is common in ML, it is usually motivated by its good performance, and not because there is any application-driven reasoning behind it; thus, there is no design transparency on this aspect. Furthermore, the reason why it works for this highly nonconvex problem is currently not well-understood from a mathematical point of view, therefore no algorithmic transparency is present.
\cite{Wu2016} introduce Physics101, a dataset which contains over 17000 video clips containing 101 objects of different characteristics, which was built for the task of deriving physical parameters such as velocity and mass.
In their work, they use the LeNet CNN architecture
to capture visual as well as physical characteristics while explicitly integrating physical laws based on material and volume to aim for scientific consistency.
Their experiments show that predictions can be made about the behavior of an object after a fall or a collision using estimated physical characteristics, which serve as input to an independent physical simulation model.
\cite{monszpart2016smash} introduce SMASH, which extracts physical collision parameters from videos of colliding objects, such as pre- and post collision velocities, to use them as input for existing physics engines for modifications.
For this, they estimate the position and orientation of objects in videos using constrained least-squares estimation in compliance with physical laws such as momentum conservation.
Based on the determined trajectories, parameters such as velocities can be derived. While their approach is based more on statistical parameter estimation than ML, their model and algorithm building process is completely transparent. Individual outcomes become explainable due to the direct relation of the computations to the underlying physical laws.

\cite{Mottaghi2016} introduce Newtonian NNs in order to predict the long-term motion of objects from a single color image. 
Instead of predicting physical parameters from the image, they introduce 12 Newtonian scenarios serving as physical abstractions, where each scenario is defined by physical parameters defining the dynamics.
The image, which contains the object of interest, is mapped to a state in one of these scenarios which best describes the current dynamics in the image.
Newtonian NNs are two parallel CNNs: one encodes the images, while the other derives convolutional filters from videos acquired with a game engine simulating each of the 12 Newtonian scenarios. The specific coupling of both CNNs in the end leads to an interpretable approach, which also (partly) allows for explaining the classification results of a single input image.

A tensor-based approach to ML for uncertainty quantification problems can be found in~\cite{schneider2018}, where the solutions to parametric convection-diffusion PDEs are learned based on a few samples. Rather than directly aiming for interpretability or explainability, this approach helps to speed up the process of gaining scientific insight by computing physically relevant quantities of interest from the solution space of the PDE. As there are convergence bounds for some cases, the design process is to some extent transparent and benefits from domain knowledge. 

An information-based ML approach using NNs to solve an inverse problem in biomechanical applications was presented in~\cite{Hoerig2017}.
Here, in mechanical property imaging of soft biological media under quasi-static loads, elasticity imaging parameters are computed from estimated stresses and strains. For a transparent design of the ML approach, domain knowledge is incorporated in two ways. First, NNs for a material model are pre-trained with stress–strain data, generated using linear-elastic equations, to avoid non-physical behavior. Second,  finite-element analysis is used to model the data acquisition process.

\vspace{1em}
\emph{Group 2b: These ML models are highly transparent, which means that the design process as well as the algorithmic components are fully accessible. The outcome of the model is explainable and the scientific consistency of the outcome is enforced.}
\vspace{1em}

For organic photovoltaics material, an approach utilizing quantum chemistry calculations and ML techniques to calibrate theoretical results to experimental data was presented by~\citep{Pyzer-Knapp2016,Lopez2017}. 
The authors consider already performed existing experiments as current knowledge, which is embedded within a probabilistic non-parametric mapping. In particular, GPs were used to learn the deviation of properties calculated by computational models from the experimental analogues. By employing the chemical Tanimoto similarity measure and building a prior based on experimental observations, design transparency is attained.
Furthermore, since the prediction results involve a confidence in each calibration point being returned, the user can be informed when the scheme is being used for systems for which it is not suited~\citep{Pyzer-Knapp2016}.
In \cite{Lopez2017}, 838 high-performing candidate molecules have been identified within the explored molecular space, due to the now possible efficient screening of over 51,000 molecules.

In \cite{Raissi2017b}, a data-driven algorithm 
for learning the coefficients of general parametric linear differential equations from noisy data was introduced, solving a so-called inverse problem.
The approach employs GP priors that are tailored to the corresponding and known type of differential operators, resulting in design and algorithmic transparency. The combination of rather generic ML models with domain knowledge in form of the structure of the underlying differential equations leads to an efficient method.
Besides classical benchmark problems with different attributes, the approach was used on an example application in functional genomics, determining the structure and dynamics of genetic networks based on real expression data. 

\vspace{1em}
\emph{Group 2c: These ML models are similar to the models in Group 2a, but besides enforced scientific consistency and plausibility of the explainable outcome, an additional post-hoc consistency check is performed.}
\vspace{1em}

In \cite{Ling2016}, a deep learning approach for Reynolds-averaged Navier–Stokes (RANS) turbulence modelling was presented. Here,  domain knowledge led to the constructions of a network architecture that embedded invariance using a higher-order multiplicative layer. This was shown to have significantly more accurate predictions compared to a generic but less interpretable NN architecture. Further, the improved prediction on a test case that had a different geometry than any of the training cases indicates that improved RANS predictions for more than just interpolation situations seem achievable. A related approach for RANS-modeled Reynolds stresses for high-speed 
flat-plate turbulent boundary layers was presented in~\cite{Wang2019}, which uses a systematic approach with basis tensor invariants proposed by~\cite{Ling2016b}. Additionally, a metric of prediction confidence and a 
nonlinear dimensionality reduction technique are employed to provide a priori assessment of the prediction confidence.

\subsubsection{Interpretation Tools for Scientific Outcomes}
Commonly used feature selection and extraction methods enhance the interpretability of the input data, and thus can lead to outcomes which can be explained by interpretable input.
Other approaches use interpretation tools to extract information from learned models and to help to scientifically explain the individual output or several outputs jointly. 
Often, approaches are undertaken to present this information via feature importance plots, visualizations of learned representations, natural language representations, or the discussion of examples. 
Nonetheless, human interaction is still required to interpret this additional information, which has to be derived ante-hoc or with help of the learned model during a post-hoc analysis.

\vspace{1em}
\emph{Group 3a: These ML approaches use interpretation tools such as feature importance plots or heatmaps to explain the outcome by means of an interpretable representation of the input. }
\vspace{1em}

While handcrafted and manually selected features are typically easier to understand, automatically determined features can reveal previously unknown scientific attributes and structures. 
\cite{ginsburg2016feature}, for example, proposes FINE (feature importance in nonlinear embeddings) for the analysis of cancer patterns in breast cancer tissue slides. This approach relates original and automatically derived features to each other by estimating the relative contributions of the original features to the reduced-dimensionality manifold.
This procedure can be combined with various, possibly intransparent, nonlinear dimensionality reduction techniques. Due to the feature contribution detection, the resulting scheme remains interpretable.

Arguably, visualizations are one of the most widely used interpretation tools. 
\cite{Hohman2018} give a survey of visual analytics in deep learning research, where such visualizations systems have been developed to support model explanation, interpretation, debugging, and improvement. The main consumers of these analytics are the model developers and users as well as non-experts. 
\cite{ghosal2018explainable} use interpretation tools for image-based plant stress phenotyping.
They train a CNN model and identify the most important feature maps in various layers that isolate the visual cues for stress and disease symptoms.
They produce so-called explanation maps as sum of the most important features maps indicated by their activation level.
A comparison of manually marked visual cues by an expert and the automatically derived explanation maps reveal a high level of agreement between the automatic approach and human ratings.
The goal of their approach is the analysis of the performance of their model, the provision of visual cues which are human-interpretable to support the prediction of the system, and a provision of important cues for the identification of plant stress.
\cite{Lerer2016} and \cite{Groth2018} use attention heatmaps to visualize the stackability of multiple wooden blocks in images. 
They conduct a conclusion study by applying localized blurring to the image and collecting the resulting changes in the stability classification of the wooden blocks into a heatmap.
Moreover, \cite{Groth2018} provide a first step towards a physics-aware model by using their trained stability predictor and heatmap analysis to provide stackability scores for unseen object sets, for the estimation of an optimal placement of blocks, and to counterbalance instabilities by placing additional objects on top of unstable stacks.

As another example, ML has been applied to functional magnetic resonance imaging data to design biomarkers that are predictive of psychiatric disorders. However, only “surrogate” labels are available, e.g., behavioral scores, and so the biomarkers themselves are also “surrogates” of the optimal descriptors~\citep{Pinho18,Varoquaux18}.
The biomarker design promotes spatially compact pixel selections, producing biomarkers for disease prediction that are focused on regions of the brain; these are then considered by expert physicians. As the analysis is based on high-dimensional linear regression approaches, transparency of the ML model is assured.
\cite{abbasi2018deeptune} introduce DeepTune, a visualization framework for CNNs, for applications in neuroscience. 
DeepTune consists of an ensemble of CNNs that learn multiple complementary representations of natural images. 
The features from these CNNs are fed into regression models to predict the firing rates of neurons in the visual cortex. 
The interpretable deepTune images, that means representative images of the visual stimuli for each neuron, are generated from an optimization process and pooling over all ensemble members.

Classical tools such as confusion matrices are also used as interpretation tools on the way to scientific outcomes. In a bio-acoustic application for the recognition of anurans using acoustic sensors, \cite{Colonna2018} use a hierarchical approach to jointly classify on three taxonomic levels, namely the family, the genus, and the species. Investigating the confusion matrix per level enabled for example the identification of bio-acoustic similarities between different species.

\vspace{1em}
\emph{Group 3b: These models are design-transparent in the sense that they use specially tailored components such as attention modules to achieve increased interpretability. The output is explained by the input using the specially selected components.}
\vspace{1em}

In \citep{Tomita19, Schlemper19} attention-based NN models are employed to classify and segment histological images, e.g., microscopic tissue images, magnetic resonance imaging (MRI), or computed tomography (CT) scans. \cite{Tomita19} found that the employed modules turned out to be very attentive to regions of pathological, cancerous tissue and non-attentive in other regions. Furthermore, \cite{Schlemper19} built an attentive gated network which gradually fitted its attention weights with respect to targeted organ boundaries in segmenting tasks. The authors also used their attention maps to employ a weakly supervised object detection algorithm, which successfully created bounding boxes for different organs.

Interpretability methods have also been used for applications which utilize time-series data, often by way of highlighting features of the sequence data. 
For example, \cite{deming2016genetic} applied attention modules in NNs trained on genomic sequences for the identification of important sequence motifs by visualizing the attention mask weights. 
They propose a genetic architect that finds a suitable network architecture by iteratively searching over various NN building blocks. 
In particular, they state that the choice of the NN architecture highly depends on the application domain, which is a challenge if no prior knowledge is available about the network design. 
It is cautioned that, depending on the optimized architecture, attention modules and expert knowledge may lead to different scientific insights. 
\cite{singh2017attend} use attention modules for genomics in their AttentiveChrome NN.
The network contains a hierarchy of attention modules to gain insights about where and what the network has focused and, thus, gaining interpretability of the results.
Also \cite{choi2016retain} developed a hierarchical attention-based interpretation tool called RETAIN (REverse Time AttentIoN) in healthcare. 
The tool identifies influential past visits of a patient as well as important clinical variables during these visits from the patient's medical history to support medical explanations.
Attention modules in recurrent NNs for multi-modal sensor-based activity recognition have been used by \cite{chen2018interpretable}. Depending on the activity, their approach provides the most contributing body parts, modals, and sensors for the network's decision.

\vspace{1em}
\emph{Group 3c: As in group 3b, these ML approaches use interpretation tools for a better understanding of the model's decision. Moreover, they integrate domain knowledge to enhance the scientific consistency and plausibility, for example, in combination with the outcome of interpretation tools.}
\vspace{1em}

\cite{kailkhura2019reliable} discusses explainable ML for scientific discoveries in material sciences. 
In their work, they propose an ensemble of simple models to predict material properties along with a novel evaluation focusing on trust by quantifying generalization performance. Moreover, their pipeline contains a rationale generator which provides decision-level interpretations for individual predictions and model-level interpretations for the whole regression model. In detail, they produce interpretations in terms of prototypes that are analyzed and explained by an expert, as well as global interpretations by estimating feature importance for material sub-classes.
Moreover, they use domain knowledge for the definition of material sub-classes and integrate it into the estimation process.
\cite{rieger2019interpretations} proposes contextual decomposition explanation penalization, which constrains a classification or regression result to more accurate and more scientifically plausible results by leveraging the explained outcome of interpretation tools. They add an explanation term in the loss function, which compares the interpretation outcome (e.g., a heatmap indicating the important parts in the image) and an interpretation provided by the user. They determine a more accurate model from an International Skin Imaging Collaboration dataset whose goal is to classify cancerous and non-cancerous images, by learning that colorful patches present only in the benign data are not relevant for classification.

\vspace{1em}
\emph{Group 3d: These approaches use the common feature-oriented representation with focus on the disentanglement of the underlying factors of variation in a system, which can be explained by an expert afterwards. Domain knowledge is employed in the design of the model and in the interpretation of the outcome.}  
\vspace{1em}

A broad framework leverages unsupervised learning approaches to learn low-complexity representations of physical process observations. 
In many cases where the underlying process features a small number of degrees of freedom, it is shown that nonlinear manifold learning algorithms are able to discern these degrees of freedoms as the component dimensions of low-dimensional nonlinear manifold embeddings, which preserve the underlying geometry of the original data space  \citep{YTCK17,DTCK18,HKBGZK18}.
It can be seen that the embedding coordinates show relation to known physical quantities. At this stage, ongoing research is focused on obtaining new scientific outcomes in new situations using this promising approach.
In a similar way, \cite{hu2017discovering} and \cite{wang2016discovering} use principal component analysis and the derived interpretable principal components for exploration of different phases, phase-transition, and crossovers in classical spin models.
Embedded feature selection schemes have been recently explored to establish or refine models in physical processes.
Using a sparsity-promoting penalty, they propose groups of variables that may explain a property of interest and promote the simplest model, i.e., the model involving the fewest variables possible while achieving a target accuracy. Domain knowledge is employed during the selection of the dictionary of candidate features.
The application of sparsity has also proved fruitful in the broader class of problems leveraging PDEs and dynamical system models~\citep{Brunton2016,MBPK16,Rudy17,SCHO13,TW17}.

The combination of parse trees with ML is investigated in~\cite{Dai2018,Li2019}. A so-called syntax-direct variational autoencoder (SD-VAE) is introduced in \cite{Dai2018}, where 
syntax and semantic constraints are used in a generative model for structured data. As an application the drug properties of molecules are predicted. The learned latent space is visually interpreted, while the diversity of the generated molecules is interpreted using domain expertise. 
The work in \cite{Li2019} uses a NN during a Monte Carlo tree search to guide its finding of an expression for symbolic regression that conforms to a set of data points and has the desired leading polynomial powers of the data. The NN learned the relation between syntactic structure and leading powers. As a proof-of-concept application, they are able to learn of physical force field, where the leading powers in the short and long ranges are known by domain experts and can used as asymptotic constraints.
\cite{Meila2018} propose a sparsity-enforcing technique to recover domain-specific meaning for the abstract embedding coordinates obtained from unsupervised nonlinear dimensionality reduction approaches in a principled fashion. The ansatz is to explain the embedding coordinates as nonlinear compositions of functions from a user-defined dictionary. 
As an illustrative example the ethanol molecule is studied, where the approach identifies the bond torsions that explain the torus obtained from the embedding method, which reflects the two rotational degrees of freedom.

\vspace{1em}
\emph{Group 3e: In addition to the works in group 3d, domain knowledge is employed to perform a-posteriori consistency checks on feature-oriented representations.}
\vspace{1em}

Feature selection schemes using embedded methods, similar to the previous group, have been used in areas such as material sciences~\citep{GVAOLDS17,Ouyang2018}. In contrast to the preceding works, additional consistency checks on the outcome of the predictive model are performed based on domain expertise, including the robustness of the model and in particular their extrapolation capability for predicting new materials.

\subsection{Scientific Outcomes by Explaining Models}
In the following examples, either interpretation tools are used to project processes in the model into a space which is interpretable or the model is designed inherently to be interpretable.
In this way, models and their components can be explained utilizing domain knowledge.

\vspace{1em}
\emph{Group 4a:  These models are designed in a transparent way and the model design enforces that model components are interpretable and scientifically explainable. Due to their design, scientific consistency and plausibility is enforced, even if not as a primary goal. The explanation of specific model components are meant to lead to novel scientific discoveries or insights.}
\vspace{1em}

Complex ML methods such as NNs, for example, can be customized to a specific scientific application so that the used architecture restricts or promotes properties that are desirable in the data modeled by the network.
For example, in \citep{Adiga2018}, an application of ML for epidemiology leverages a networked dynamical system model for contagion dynamics, where nodes correspond to subjects with assigned states; thus, most properties of the ML model match the properties of the scientific domain considered. 
A complex NN is reduced by \cite{tanaka2019deep} to understand processes in neuroscience.
By reducing the number of units in the complex model by means of a quantified importance utilizing gradients and activation values, a simple NN with one hidden layer is derived which can be easily related to neuroscientific processes.
\cite{Lusch2018} construct a NN for computing Koopman eigenfunctions from data. Motivated by domain knowledge, they employ an auxiliary network to parameterize the continuous frequency. Thereby, a compact autoencoder model is obtained, which additionally is interpretable. For the example of the nonlinear pendulum, the two eigenfunctions are learned with a NN and can be mapped into magnitude and phase coordinates. In this interpretable form, it can be observed that the magnitude traces level sets of the Hamiltonian energy, a new insight which turned out to be consistent with recent theoretical derivations beforehand unknown to the authors.
\cite{Ma2018} introduces visible NNs, which encode the hierarchical structure of a gene ontology tree into an NN, either from literature or inferred from large-scale molecular data sets. This enables transparent biological interpretation, while successfully predicting effects of gene mutations on cell proliferation. Furthermore, it is argued that the employed deep hierarchical structure captures many different clusters of features at multiple scales and  pushes interpretation from the model input to internal features representing biological subsystems.
In their work, despite no information about subsystem states was provided during model training, previously undocumented learned subsystem states could be confirmed by molecular measurements.

Beside NNs, also other ML algorithms can be used to derive scientific outcomes from an interpretable model.
\cite{daniels2019automated} use their ML algorithm `Sir Isaac' %\citep{daniels2015automated} 
to infer a dynamical model of biological time-series data to understand and predict dynamics of worm escape behavior.
They model a system of differential equations, where the number of hidden variables is determined automatically from the system, and their meaning can be explained by an expert.

\cite{Iten2018} introduces SciNet, a modified variational autoencoder which learns a representation from experimental data and uses the learned representation to derive physical concepts from it rather than from the experimental input data.
The learned representation is forced to be much simpler than the experimental data, for example by being captured in a few neurons, and contains the explanatory factors of the system such as the physical parameters. 
This is proven by the fact that physical parameters and the activations of the neurons in the hidden layers have a linear relationship.
Additionally, \cite{ye2018interpretable} construct the bottleneck layer in their NN to represent physical parameters to predict the outcome of a collision of objects from videos. 
However, the architecture of the bottleneck layer is not learned, but designed with prior knowledge about the underlying physical process.

Understanding structures such as groups, relations, and interactions is one of the main goals to achieve scientific outcomes. However, it constitutes a core challenge and so far only limited amount of work has been conducted in this area.
\cite{yan2019groupinn}, for example, 
introduce a grouping layer in a graph-based NN called GroupINN to identify subgroups of neurons in an end-to-end model. 
In their work, they build a network for the analysis of time-series of functional magnetic resonance images of the brain, which are represented as functional graphs, with the goal to reveal relationships between highly predictive brain regions and cognitive functions. 
Instead of working with the whole functional graph, they exploit a grouping layer in the network to identify groups of neurons, where each neuron represents a node in the graph and corresponds to a physical region of interest in the brain.
The grouped nodes in the coarsened graph are assigned to regions of interest, which are useful for prediction of cognitive functions, and the connections between the groups are defined as functional connections.

\cite{tsang2017detecting} introduces neural interaction detection, a framework with variants of feedforward NNs for detecting statistical interactions. By examining the learned weight matrices of the hidden units, their framework was able to analyze feature interactions in a Higgs boson dataset. Specifically, they analyze feature interactions in simulated particle environments which originate from the decay of a Higgs boson.
Deep tensor networks are used by \cite{schutt2017quantum} in quantum chemistry to predict molecular energy up to chemical accuracy, while allowing interpretations. A so-called local chemical potential, a variant of sensitivity analysis where one measures the effect on the NN output of inserting a charge at a given location, can be used to gain further chemical insight from the learned model. As an example, a classification of aromatic rings with respect to their stability can be determined from these three-dimensional response maps.

\vspace{1em}
\emph{Group 4b: These ML models are designed with a high transparency and with the goal to derive scientifically plausible results. Due to this, the outcome of the model and the model components themselves are interpretable and can be scientifically explained. 
In contrast to the works presented in group 4a, the following examples deal with also algorithmically transparent methods.}
\vspace{1em}

Different types of physics-aware GP models in remote sensing were studied by~\cite{CampsValls2018} with the goal to estimate bio-physical parameters such as leaf area index.
In one case, a latent force model that incorporates ordinary differential equations is used in inverse modelling from real in-situ data. 
The learned latent representation allowed an interpretation in view of the physical mechanism that generated the input-output observed relations, i.e., one latent function captured the smooth and periodic component of the output, while two other focus on the noisier part with an important residual periodical component.
So-called order parameters in condensed matter physics are analysed in~\citep{Greitemann2019,Liu2019}. Using domain knowledge, a kernel is introduced to investigate $O(3)$-breaking orientational order. A two-class and a multi-class setting are tackled with support vector machines (SVM). The decision function is physically interpreted as an observable corresponding to an order parameter curve, while the bias-term of the SVM can be exploited to detect phase transitions. Furthermore, nontrivial blocks of the SVM kernel matrices can be identified with so-called spin color indices. In these works, for spin and orbital systems the analytical order parameters could be extracted.

\subsection{Related Surveys about Machine Learning in the Natural Sciences}

\cite{Butler2018} give on overview on recent research using ML for molecular and materials science. Given that standard ML models are numerical, the algorithms need suitable numerical representations that capture relevant chemical properties, such as the Coulomb matrix and graphs for molecules, and radial distribution functions that represent crystal structures. Supervised learning systems are in common use to predict numerical properties of chemical compounds and materials. Unsupervised learning and generative models are being used to guide chemical synthesis and compound discovery processes, where deep learning algorithms and generative adversarial networks have been successfully employed. Alternative models exploiting the similarities between organic chemistry and linguistics are based on textual representations of chemical compounds.

A review of the manifold recent research in the physical sciences is given by \cite{RevModPhys.91.045002}, with applications in particle physics and cosmology, quantum many-body physics, quantum computing, and chemical and material physics. The authors observe a surge of interest in ML, while noting that the research is starting to move from exploratory efforts on toy models to the use of real experimental data. It is stressed that an understanding of the potential and the limitations of ML includes an insight into the breaking point of these methods, but also the theoretical justification of the performance in specific situations, be it positive or negative.

In single-cell genomics, computational data-driven analysis methods are employed to reveal the diverse simultaneous facets of a cell's identity, including a specific state on a developmental trajectory, the cell cycle, or a spatial context. The analysis goal is to obtain an interpretable representation of the dynamic transitions a cell undergoes that allows to determine different aspects of cellular organization and function. There is an emphasis on unsupervised learning approaches to cluster cells from single-cell profiles, and thereby to systematically detect beforehand unknown cellular subtypes, for which then defining markers are investigated in a second step, see~\citep{Wagner2016} for a review on key questions, progress, and open challenges in this application field.

Several ML approaches have been used in biology and medicine to derive new insights, as described in \cite{ching2018opportunities} for the broad class of deep learning methods. Supervised learning mostly focuses on the classification of diseases and disease types, patient categorization, and drug interaction prediction. Unsupervised learning has been applied to drug discovery.
The authors point out that in addition to the derivation of new findings, an explanation of these is of great importance. Furthermore, the need in deep learning for large training datasets poses a limit to its current applicability beyond imaging (through data augmentation) and so-called `omics' studies. An overview of deep learning approaches in systems biology is given in \cite{Gazestani2019}. They describe how one can design NNs  that encode the extensive, existing network- and systems-level knowledge that is generated by combing diverse data types. It is said that such designs inform the model on aspects of the hierarchical interactions in the biological systems that are important for making accurate predictions but are not available in the input data.
\cite{holzinger2019causability} discuss the difference between explainability and causality for medical applications, and the necessity of a person to be involved.
For the successful application of ML for drug design, \cite{Schneider2019} identify five `grand challenges': obtaining appropriate datasets, generating new hypotheses, optimizing in a multi-objective manner, reducing cycle times, and changing the research culture and mindset. These underlying themes should be valid for many scientific endeavours. 

\cite{reichstein2019deep} give an overview of ML research in Earth system science. They conclude that while the general cycle of exploration and hypotheses generation and testing remains the same, modern data-driven science and ML can extract patterns in observational data to challenge complex theories and Earth system models, and thereby strongly complement and enrich geoscientific research.
Also \cite{Karpatne2018a} point out that a close collaboration with domain experts in the geoscientific area and ML researchers is necessary to solve novel and relevant tasks. They state that developing interpretable and transparent methods is one of the major goals to understand patterns and structures in the data and to turn it into scientific value. 

\section{Discussion}

In this work, we reviewed the concept of explainable machine learning and discerned between transparency, interpretability and explainability. We also discussed the possibility of influencing model design choices and the step of interpreting algorithmic outputs by domain knowledge and a posteriori consistency checks. We presented a more fine-grained characterization of different stages of explainability, which we briefly elaborated on by means of several recent exemplary works in the field of machine learning in the natural sciences.

While machine learning is employed in uncountable scientific projects and publications nowadays, the vast majority is not concerned with aspects of interpretability or explainability. We argue that the latter is a crucial part for extracting truly novel scientific results and ideas from employing ML methods. Therefore, we hope that this survey provides new ideas and methodologies to scientists looking for means to explain their algorithmic results or to extract relevant insights on the corresponding study object.

Finally, note that as an additional component in the scientific data analysis workflow of the future, we expect causal inference~\citep{Pearl2011,Scholkopf2019} to play a role. Having said this, we believe that causal inference will reguire even more basic research than what is still needed for the uptake of explainable machine learning in the natural sciences.

\section*{Acknowledgements}
Part of the work was performed during the long-term program on ``Science at Extreme Scales: Where Big Data Meets Large-Scale Computing'' held at the Institute for Pure and Applied Mathematics, University of California Los Angeles, USA.
We are grateful for their financial support during the program. We cordially thank the participants of the long term program for fruitful discussions, in particular Keiko Dow, Longfei Gao, Pietro Grandinetti, Philipp Haehnel,
Mojtaba Haghighatlari, and René Jäkel.

\bibliographystyle{plainnat}
\bibliography{references}
\end{document}